\documentclass[lettersize,journal]{IEEEtran}
\usepackage{amsmath,amsfonts}
\usepackage{algorithmic}
\usepackage{algorithm}
\usepackage{array}
\usepackage[caption=false,font=normalsize,labelfont=sf,textfont=sf]{subfig}
\usepackage{textcomp}
\usepackage{stfloats}
\usepackage{url}
\usepackage{pgf}
\usepackage{verbatim}
\usepackage{graphicx}
\usepackage{booktabs}
\usepackage{cite}
\usepackage{xcolor}
\usepackage{threeparttable}
\usepackage{authblk}

\begin{document}

\def\mathdefault#1{#1}

\title{ADAPTS: Agentic Decomposition for Automated Protocol-agnostic Tracking of Symptoms}
\author{
\IEEEauthorblockN{Alexandria K. Vail, Marcelo Cicconet, Katie Aafjes-van Doorn, Ryan Maroney, Marc Aafjes} \\
\IEEEauthorblockA{Deliberate Solutions, Inc. \\
New York, New York, USA}}

\maketitle

\setcounter{dbltopnumber}{1}

\begin{abstract}
Modeling latent clinical constructs from unconstrained clinical interactions is a unique challenge in affective computing. We present ADAPTS (Agentic Decomposition for Automated Protocol-agnostic Tracking of Symptoms), a framework for automated rating of depression and anxiety severity using a mixture-of-agents LLM architecture. This approach decomposes long-form clinical interviews into symptom-specific reasoning tasks, producing auditable justifications while preserving temporal and speaker alignment. Generalization was evaluated across two independent datasets ($N=204$) with distinct interview structures. On high-discrepancy interviews, automated ratings approximated expert benchmarks ($\text{absolute error}=22$) more closely than original human ratings ($\text{absolute error}=26$). Implementing an ``extended'' protocol that incorporates qualitative clinical conventions significantly stabilized ratings, with absolute agreement reaching $\text{ICC(2,1)} = 0.877$. These findings suggest that the ADAPTS framework enables promising evaluations of psychiatric severity. While the current implementation is purely text-based, the underlying architecture is readily extensible to multimodal inputs, including acoustic and visual features. By approximating expert-level precision in a protocol-agnostic manner, this framework provides a foundation for objective and scalable psychiatric assessment, especially in resource-limited settings.
\end{abstract}


\section{Introduction}

Depression and anxiety are among the most prevalent mental health conditions worldwide, and accurate, timely assessment is essential for effective treatment and evaluation of intervention efficacy. In both clinical practice and clinical trials, clinician-administered interviews—such as the Hamilton Depression Rating Scale (HAM-D;~\cite{hamilton1960rating}), the Hamilton Anxiety Rating Scale (HAM-A;~\cite{hamilton1959assessment}), and the Montgomery–Åsberg Depression Rating Scale (MADRS;~\cite{montgomery1979new}) --- are widely regarded as the gold standard for measuring symptom severity and change over time. These instruments were designed to capture nuanced clinical judgment by integrating patient self-report with clinician observation, and they remain the primary endpoints by which regulatory bodies such as the U.S. Food and Drug Administration evaluate antidepressant efficacy~\cite{fda2018major}.

Despite their central role, clinician-administered interviews are costly, time-intensive, and difficult to scale, requiring extensive training and ongoing calibration to maintain reliability. In routine clinical practice, structured rating scales are rarely used, often due to time constraints and concerns about feasibility~\cite{demyttenaere2020trends}. Moreover, even under controlled conditions, clinician ratings exhibit substantial interrater variability, reflecting both subjective judgment and rater-specific thresholds~\cite{kobak2004rater}. This measurement noise inflates variance, reduces statistical power, and has been implicated as a contributor to failed late-stage clinical trials in psychiatry~\cite{muller2002effects}. While self-report scales offer efficiency, they frequently lack the diagnostic rigor of clinician-administered assessments and are more susceptible to patient-specific biases; meta-analytic work shows that clinician-rated scales often yield larger and more reliable effect sizes than self-report measures within the same trials~\cite{cuijpers2010selfreported}. Clinical ratings are therefore noisy even at the benchmark level~\cite{kobak2004rater}, and this noise directly impacts statistical conclusions and downstream clinical decision-making.

Recent advances in affective computing have enabled machines to infer human emotional and psychological states from speech, text, and multimodal interaction signals. Instruction-tuned large language models (LLMs) demonstrate strong zero-shot reasoning capabilities over long-form text, enabling them to analyze complex narratives without task-specific fine-tuning. In mental health research, LLMs have been explored for symptom detection, diagnostic inference, and clinical documentation analysis~\cite{raganatoleveraging}, and recent work suggests that LLM-based ratings can approach human-level agreement under certain conditions~\cite{kebe2025llamadrs,weber2025using}. However, translating these capabilities into clinically meaningful severity assessment—particularly from naturalistic clinician–patient dialogue—remains difficult. Clinical interviews are long, multi-speaker, and semi-structured; symptom-relevant evidence is expressed through emotionally nuanced dialogue with interruptions, overlapping speech, and clinically meaningful content scattered throughout the session and interleaved with clinician prompts. For ``AI that cares'' in practice, systems must generate outputs in the language of clinical decision-making (e.g., symptom severity ratings) while remaining auditable and calibrated, not merely correlated with clinician scores.

Two technical gaps limit many existing LLM-based approaches in this setting. First, many systems treat assessment as a monolithic prediction task and produce a single global score without preserving item-level reasoning or auditability. End-to-end pipelines that collapse interviews into global embeddings risk context dilution, obscuring localized affective cues that clinicians rely on when rating specific symptom domains. Second, clinical assessment protocols vary substantially across settings: some interviews administer scales sequentially, while others collapse overlapping items into unified flows to reduce burden. Methods that assume a fixed question–answer order are therefore brittle, failing silently when applied to heterogeneous protocols.

Moreover, LLM reliability over long contexts is imperfect. Empirical work shows that even long-context models can fail to use relevant information consistently when it appears in the middle of long inputs (``lost in the middle'')~\cite{liu2024lost}, motivating designs that reduce reliance on monolithic, single-pass reasoning. This motivates architectural approaches that decompose long clinical interviews into modular, symptom-specific reasoning tasks, mirroring how human clinicians evaluate individual items rather than forming impressions holistically. Modular agent-based approaches, combined with explicit justification and anchored scoring logic, offer a principled pathway toward transparent, reliable affective inference.

This motivates a specific technical thesis: clinical severity rating from interviews should be treated as a decomposable evidence-retrieval and calibration problem rather than a single global prediction problem. Concretely, systems should (i) retrieve symptom-specific evidence from anywhere in the interview, (ii) maintain speaker-aware, time-aligned structure to avoid attribution errors, (iii) generate auditable justifications, and (iv) evaluate performance using metrics that separate rank-order reliability from absolute calibration and bias [21]. This framing is not merely methodological—models may reproduce relative severity patterns while still exhibiting systematic point offsets, indicating stable but miscalibrated thresholds rather than catastrophic instability.

In this work, we address these gaps by introducing ADAPTS (Agentic Decomposition for Automated Protocol-agnostic Tracking of Symptoms), modular LLM framework for automated re-rating of depression and anxiety severity from authentic clinical interview audio. The ADAPTS framework combines speaker-aware, temporally aligned preprocessing with item-level sub-agents that retrieve symptom evidence across the entire interview, produce item-level justifications, and map evidence to anchored severity scores. We evaluate the system across heterogeneous clinical protocols in two independent datasets and adopt a multi-metric evaluation strategy that explicitly distinguishes rank-order consistency from calibration bias. By grounding automated affective inference in gold-standard clinical instruments and transparent reasoning, this study demonstrates how LLMs can support scalable, reliable, and interpretable mental health measurement.
\section{Related Work}

\subsection{Affective computing and mental health inference from language and speech}
Affective computing has long studied inference of latent affective and psychological states from speech and text; mental health applications are a major extension of this agenda. Contemporary surveys and scoping reviews highlight increasing use of LLMs in mental health care but also recurrent concerns regarding evaluation quality, safety, bias, and the difficulty of grounding outputs in clinically verifiable evidence~\cite{weber2025using}. In parallel, NLP work has moved from social media detection toward more clinically proximal sources, including patient narratives and clinical interactions, to improve ecological validity—though clinical dialogue remains more complex than typical text classification settings.
Affective computing research on speech has also made major progress in recognizing emotion and related paralinguistic phenomena (e.g., arousal, valence, stress markers), supported by survey work and benchmark-driven progress in speech emotion recognition. Within mental health, a parallel line of work has focused on automatic depression and distress detection from interview data using audio, text, or multimodal features. Community benchmarks such as the Audio/Visual Emotion Challenge (AVEC) introduced shared tasks on depression-related inference and fostered comparability across methods~\cite{valstar2013avec}, while clinical interview corpora such as the Distress Analysis Interview Corpus (DAIC) and its DAIC-WOZ subset have supported distress inference from human–agent and clinician–patient interviews~\cite{devault2014simsensei}.

\subsection{LLMs for structured symptom severity scoring from interviews}

A rapidly emerging line of work applies LLMs to score structured psychiatric instruments directly from interview transcripts. Prompt-engineered LLMs have been used to score clinician-administered MADRS interviews with high correlations to human ratings~\cite{raganatoleveraging}. Notably, work by Kebe et al. has demonstrated that prompting open-source LLMs can approximate clinician scoring for interview-based depression assessment using MADRS, reporting strong agreement and highlighting the feasibility of item-level prompting strategies on real interviews~\cite{kebe2025llamadrs}. Separately, Weber et al.~\cite{weber2025using} report symptom-based depression evaluation using a fine-tuned language model, supporting the idea that adaptation to clinical data can improve prediction of symptom-level outcomes. However, these studies generally do not address protocol robustness—that is, whether systems generalize across different interview structures—and often emphasize association metrics without fully characterizing calibration error or systematic bias. This leaves open important questions about robustness across protocols and instruments and underscores the need to report both concordance and absolute error.

\subsection{Long-context behavior, modularity, and agentic strategies}

LLMs’ long-context behavior is an active research area that is directly pertinent to interview-based mental health assessment. Evidence shows that position effects can degrade retrieval and task performance even in models designed for long contexts~\cite{liu2024lost}. This motivates decomposed or agentic strategies that interleave retrieval and reasoning to improve robustness and interpretability. For example, ReAct-style paradigms explicitly combine reasoning traces with actions such as evidence gathering, improving interpretability and reducing error propagation in complex tasks~\cite{yao2022react}. In the clinical interview setting, modular symptom agents provide a natural analog: each agent performs targeted retrieval of symptom-relevant evidence and produces localized justifications, reducing context dilution and supporting audit trails that align with clinical reasoning practices.

\subsection{Reliability, calibration, and evaluation beyond correlation}

Evaluation of automated clinical assessment systems must extend beyond simple association metrics. High correlations can coexist with systematic bias—particularly when models learn consistent but shifted severity thresholds. Reliability metrics such as intraclass correlation coefficients (ICC) are widely used to assess agreement and consistency in clinical rating contexts, and clear guidelines exist for selecting and reporting ICC appropriately~\cite{koo2016guideline}. For systems intended to support care or clinical research, it is equally important to quantify absolute error (e.g., MAE, RMSE) and to test for directional bias in residuals. This distinction matters: stable, systematic offsets may be addressable through calibration layers, whereas large outliers may signal hallucinations or brittle reasoning that undermines clinical safety and utility.

\subsection{Preprocessing foundations: transcription, alignment, diarization, and protocol heterogeneity}

For interview-based mental health inference, upstream errors in transcription, temporal alignment, or speaker attribution can cascade into downstream scoring errors. Automated inference from clinical interviews introduces technical barriers that are less prominent in typical affect recognition settings: interviews are long and information-dense, clinically relevant evidence may appear far from the probes that elicited it, speech can overlap, and turn-taking may be irregular. For clinically meaningful inference, it is often essential to align a clinician’s probe with a patient’s behavioral response in time --- both to locate evidence and to support explainability. Modern automatic speech recognition and alignment tools provide a pathway to structured, temporally anchored transcripts. Whisper, trained via large-scale weak supervision, demonstrated strong zero-shot robustness across diverse audio conditions~\cite{radford2023robust}. WhisperX extended this foundation with word-level time alignment and long-form transcription, addressing timestamp drift and improving temporal reliability through forced phoneme alignment~\cite{bain2023whisperx}. For speaker diarization --- the ``who spoke when'' problem --- pyannote.audio provides neural building blocks and pretrained pipelines widely used in contemporary diarization systems~\cite{bredin2023pyannote}. Together, these tools enable the construction of diarized, time-aligned corpora that are prerequisites for reliable affective inference in multi-speaker clinical dialogue.

Another key but under-discussed obstacle for clinical affective AI is protocol heterogeneity. Many mental health datasets have relatively stable interview structures, which can induce models to exploit ordering artifacts or interviewer-driven regularities. Recent work has raised concerns about how interviewer prompts can bias depression detection and limit generalization when prompts are used as features rather than as contextual scaffolding~\cite{burdisso2024daicwoz}. In real clinical practice, protocol structure varies: scales may be administered sequentially, interwoven, or collapsed; probes may recur; and symptom evidence relevant to one construct may appear during discussion of another. Robust systems should therefore be protocol-agnostic, retrieving symptom-relevant evidence across the interview rather than assuming a linear questionnaire flow

\section{Data Acquisition and Preprocessing}

The primary objective of the preprocessing phase is to convert unstructured clinical audio into a structured, speaker-aware format that preserves any symptomatic nuances of the patient's speech. The technical challenges in automatic assessment lie in the prevalence of overlapping speech and the requirement for precise temporal alignment between a clinician's diagnostic query and the patient's behavioral response.

The generalizability of our automated framework was evaluated using two independent datasets representing two distinct protocol structures. The first dataset, designated as ``DHRI'' ($n=121$), consists of interviews with participants recruited from outpatient eating disorder services, where most patients reported mild depressive symptoms. The interview protocol utilizes a customized clinical protocol that integrates the Hamilton Depression Rating Scale (HAM-D;~\cite{hamilton1960rating}), the Hamilton Anxiety Rating Scale (HAM-A;~\cite{hamilton1959assessment}), and the Montgomery-\AA{}sberg Depression Rating Scale (MADRS;~\cite{montgomery1979new}) into a single, non-redundant interview flow, based on the HAM-D/MADRS Interview protocol pioneered by Iannuzzo et al.~\cite{iannuzzo2006development}. This structure minimizes rater fatigue and participant burden by collapsing overlapping questionnaire items into a unified stream. Consequently, the resulting dialogue does not follow the linear progression of a single instrument.

In contrast, the ``Illiad'' dataset ($n=83$), recruited from a depression research study, adheres to a traditional sequential protocol, in which the standard HAM-D protocol is administered in its entirety, followed immediately by the HAM-A protocol in its entirety. To accommodate these structural variations, the re-rating agents were engineered to be protocol agnostic: rather than relying on a fixed, linear question-and-answer order, the agents utilize an agnostic retrieval procedure, scanning the entire transcribed interview to identify relevant symptomatic content regardless of its chronological position in the interview. This approach ensures that information elicited during a depression probe that might be relevant to an anxiety rating is not overlooked by the modular architecture due to chronological constraints.

\subsection{Transcription and Diarization}

The transcription of raw audio recordings into structured text followed a four-step pipeline to ensure temporal and linguistic precision in noisy naturalistic dialog: Voice Activity Detection (VAD), large-scale transcription, phoneme-level alignment, and speaker diarization.

The pipeline begins with a Voice Activity Detection (VAD) pass using the WhisperX framework~\cite{bain2023whisperx}, segmenting the audio stream and filtering ambient acoustic noise or prolonged silence. Particularly in psychiatric contexts, where patients may exhibit significant latencies in speech due to psychomotor retardation or high cognitive load~\cite{sobin1997psychomotor}, VAD mitigates the risk of the transcription engine generating spurious text (``hallucinations'') during non-speech intervals~\cite{baranski2025investigation}. Following the VAD pass, the speech segments are transcribed using the Whisper \texttt{large-v2} model, selected for its robust performance in handling the varied of prosodic and emotional vocalizations typical of clinical encounters~\cite{radford2023robust}.

Another significant challenge in automatic speech recognition (ASR) is the phenomenon of timestamp drift, which can uncouple the transcribed text from the speaker's nonverbal or other behaviors. To mitigate this risk, the framework implements a word-level alignment pass using a language-specific phoneme model: for the English language, this defaults to a Wav2Vec2-based model~\cite{bredin2020pyannoteaudio,wang2020fairseq}. This granularity anchors symptomatic reports to the original audio timeline, providing a reliable foundation for future multimodal analyses correlating verbal reports with non-verbal behavioral cues.

The final stage of the pipeline is a speaker diarization pass using Pyannote-based logic~\cite{bain2023whisperx} integrated within the WhisperX library. This process identifies unique speaker embeddings to partition the transcript into discrete speaker turns. However, because diarization labels are inherently anonymous (e.g., \texttt{SPEAKER\_01}), our framework applies a procedural heuristic for role attribution. Specifically, the system utilizes a lexical density and interrogative check: the speaker initiating the interview with interrogative syntax matching standardized Hamilton interview probes is assigned the clinician label, while the other speaker is assigned the patient label. The finalized output is a temporally-aligned, diarized dialog corpus.

\subsection{Ground Truth Calibration and Expert Validation}

Establishing a robust ``ground truth'' benchmark is a methodological requirement for interpreting the system's performance, particularly given the inherent subjectivity of human psychiatric assessment. To mitigate rater-induced variance, this study implemented a rigorous multi-stage calibration process.

All interviews and initial ratings were conducted by a diverse panel of clinical professionals, including licensed clinical psychologists, licensed social workers, and doctoral-level graduate students. Before data collection, each rater underwent standardized training for the administration and scoring of Hamilton protocols, to achieve a minimum intraclass correlation coefficient (ICC) of 0.8, frequently cited as indicative of good inter-rater reliability in clinical research~\cite{koo2016guideline}. To prevent ``rater drift'' --- in which a clinician's scoring criteria shift longitudinally due to habituation or fatigue --- all raters participated in weekly continued calibration and quality checks~\cite{williams1988structured}.

The methodology for defining the ground truth differed slightly between the two datasets due to their distinct protocol structures: DHRI utilized the median score of two or more raters to filter out individual bias, while Illiad initially relied on single human ratings. To address the increased subjectivity of single-rater metrics, we recruited a senior expert clinical rater with 13 years of experience of central rating quality reviews to blindly re-rate recordings of high-discrepancy cases. 

This validation procedure yielded significant insight into the system's performance. When evaluating the interviews with the highest initial discrepancy, the Sum of Absolute Errors (SAE) was calculated on the total score of HAM-D (which ranges from 0 to 52). AI-generated ratings were significantly closer to the expert's benchmarks (cumulative SAE of 22 points) than the original human ratings were to the same expert (cumulative SAE of 26 points). These findings suggest that the automated agent may function not only as a proxy but also as a corrective layer capable of identifying and stabilizing outliers in human clinical ratings.

\subsection{The Extended HAM-D Protocol}

To evaluate the impact of explicit clinical knowledge injection, we also evaluated an ``Extended'' HAM-D 17* variant: while the standard GRID-HAMD simply provides guidelines for score assignment based on a fixed frequency and intensity mapping, this protocol augments agent prompts with item level qualitative conventions derived from expert clinical notes. These conventions provide explicit logic for handling edge cases, such as excluding insomnia caused by unambiguous external factors, or distinguish between realistic self-reproach and pathological guilt.
\section{The ADAPTS Framework}

The architectural design of the proposed framework prioritizes modularity, to replicate the item-level scrutiny traditionally applied by trained human clinicians. One challenge in utilizing Large Language Models (LLMs) for psychiatric assessment is the risk of ``context dilution'' --- in which the model loses focus on specific symptomatic details when forced to process an entire hour-long clinical encounter in a single pass. To mitigate this, rather than employing a monolithic model to generate a global score, ADAPTS utilizes a distributed mixture-of-agents approach: a decentralized network of specialized agents, each engineered to perform a discrete, well-defined task.

The system achieves this through the functional decomposition of our established psychometric instruments. For the HAM-D 17, our framework employs 15 independent sub-agents, each responsible for evaluating a single symptomatic category, (e.g., Depressed Mood, Insomnia, or Suicidal Ideation). A parallel structure of 13 sub-agents was used for the HAM-A 14 anxiety assessment. Each agent therefore functions as a dedicated rater for its assigned symptom, scanning the diarized transcript for relevant evidence and providing a localized justification before rendering a quantitative score.

To ensure clinical validity, the framework is intentionally constrained to items that can be substantiated through text. Consequently, items that require direct physical or paralinguistic observation were excluded from the automated rating process. These omissions include the HAM-D items for Psychomotor Retardation and Psychomotor Agitation (items 8 and 9), and the HAM-A item for Overall Anxious Behavior (item 14). The rationale for these exclusions is that the recognition of such symptoms relies heavily upon visual and motor cues, such as physical restlessness, slowed movement, or facial expression, that are not reliably captured from transcribed text. Excluding these observational items provides a methodological safeguard to prevent the model from attempting to infer behaviors beyond the scope of the available textual data.

\subsection{Agent Selection and Model Benchmarking}

The selection of LLMs for this framework was predicated on achieving a representative cross-section of current state-of-the-art architectures, varying in both parameter scale and training objectives. We benchmarked our framework across five models to evaluate how differing optimization strategies, such as proprietary instruction tuning versus open-weights reasoning focus, impact clinical rating reliability. The framework's performance was measured against these distinct architectural classes in order to determine if effective clinical calibration is a function of raw parameter scale or whether reasoning-specific tuning is truly required for true affective alignment.

\begin{itemize}
    \item \emph{Claude Sonnet 4.5 and Gemini 3 Pro:} These proprietary models were selected for their documented proficiency in long-context processing and zero-shot reasoning. Given that clinical interviews may often exceed 60 minutes of dialog, these models provide a high-fidelity baseline for potential ``lost in the middle'' retrieval failures.

    \item \emph{DeepSeek R1:} This model was included to evaluate the efficacy of specialized ``reasoning'' models that utilize large-scale reinforcement learning to emphasize chain-of-thought processing, aligning with our modular requirement for generating explicit qualitative justification before quantitative mapping is performed.

    \item \emph{Llama Scout 4:} As a leading open-weights model, Llama Scout represents the current ceiling for non-proprietary instruction tuning. Its inclusion allows for an assessment of whether modular, item-level prompting can enable open-source architectures to approximate the performance of larger, closed-source systems.

    \item \emph{GPT OSS:} A localized open-source GPT variant was also included to establish a performance floor for standard transformer-based inference without the specialized clinical or reasoning-specific tuning found in the higher-tier benchmarks.
\end{itemize}

\subsection{Prompt Engineering and GRID Structure}

The internal logic governing each sub-agent is structured to ensure that qualitative reasoning is explicitly enumerated before a final quantification is suggested. This paradigm provides clinicians with a verifiable audit trail and improves consistency by enforcing a chain-of-thought inference process~\cite{wei2022chainofthought}.

The first instruction is \emph{contextual identification}, where the agent is instructed to scan the complete diarized interview transcript to isolate specific dialog relevant to its assigned symptomatic focus. Once the relevant context is isolated, the agent is instructed to generate a \emph{qualitative justification}, citing specific evidence from the patient's speech, such as specific phrases or reported behaviors,  to support its quantitative score. This step facilitates clinical explainability, allowing for human-in-the-loop auditing of the AI's decision-making process. The final stage involves the \emph{quantitative mapping} from the qualitative summary into a standardized integer (typically 0--4), with logic varying by instrument.

\begin{itemize}
    \item \emph{HAM-D (GRID-HAMD Structure):} For the assessment of depressive symptoms, the framework incorporates the GRID-HAMD structure~\cite{williams2008gridhamd}. This variant of the Hamilton scale requires the agent to evaluate symptoms across two dimensions: frequency and intensity. By requiring separate assessments for how often a symptom occurs and the severity of its manifestation, the GRID-HAMD structure minimizes the risk of overrating a symptom that is discussed at length but lacks clinical severity, or underrating a severe symptom that is only briefly mentioned.
    \item \emph{HAM-A (Severity Anchor Structure):} In contrast, for anxiety symptoms, the sub-agents map qualitative evidence directly to a singular 0--4 severity scale. Notably, a standardized grid-equivalent structure for the HAM-A does not presently exist in the clinical literature. Instead, the rating guided by specific clinical anchor points (e.g., Mild, Moderate, Severe) defined by the standard HAM-A manual~\cite{hamilton1959assessment}. The agent is tasked with synthesizing the reported intensity of psychological or somatic anxiety into a single representative integer, following a direct mapping process rather than the dual-axis reconciliation required for the HAM-D.
\end{itemize}
\section{Experimental Evaluation}

To evaluate the performance of our modular rating framework, we established a multi-metric statistical evaluation protocol, distinguishing between the statistical relationship of the ratings (concordance and association) and the absolute magnitude of the discrepancies (error analysis).

\subsection{Concordance and Association}

This dimension evaluates how reliably the LLM agent can replicate the assessment patterns of human-expert benchmarks, measuring the stability and strength of the relationship across clinical score distributions. We utilize Pearson's $r$ and Spearman's $\rho$: while Pearson's $r$ measures linear strength, Spearman's $\rho$ measures the monotonic relationship based on rank order~\cite{schober2018correlation,hauke2011comparison}. High correlation suggests that the agent can internalize the ordinal structure of the symptom severity scale, prioritizing high-severity patients even if the absolute point values are slightly shifted.

While linear and rank correlation metrics evaluate whether the agent can accurately predict clinical variance, they are still insensitive to systematic bias in the scoring scale. To address this limitation, we also utilize the Intraclass Correlation Coefficient (ICC) as a measure of reliability. The ICC evaluates reliability by comparing the variance between participants with the overall variance, including rater-introduced variance. We report ICC(3,1) (measuring consistency) to evaluate relative ranking stability, and ICC(2,1) (measuring absolute agreement) to determine the extent to which the LLM agent's scores accurately approximate the expert-level scores. High consistency is vital for tracking longitudinal patient trajectories (the ``delta''), ensuring that any reported changes reflect actual clinical progress as opposed to stochastic noise.

\subsection{Error Analysis}

This dimension evaluates how reliably the LLM agent can approximate the absolute values of expert ratings, measuring the precision of individual scores and identifying systematic bias. We quantify interview-level discrepancy using Mean Absolute Error (MAE) and Root Mean Square Error (RMSE). MAE represents the ``typical'' error magnitude, while RMSE weights larger errors more heavily to penalize catastrophic misinterpretation or spurious output (e.g., LLM hallucinations).

Given that clinical ratings will often exhibit non-normal distributions or floor effects, we apply the Wilcoxon signed-rank test~\cite{wilcoxon1945individual} to the raw residuals (the difference between the agent's score and the expert score). This non-parametric approach determines if the agent's ratings exhibit systematic bias without assuming a normal distribution of residuals. A significant result suggests a systematic directional flaw, e.g., an agent consistently over- or under-estimating symptom severity. For clarity in presentation, we report these test results in our tables by annotating the MAE of each item with its corresponding $p$-value significance level.

To maintain the statistical integrity of these metrics across multi-item scales like the HAM-D and HAM-A, we apply the Benjamini-Hochberg~\cite{benjamini1995controlling} procedure to control the false discovery rate. This correction mitigates the risk that any significant results are simply the product of statistical noise inherent in multiple comparisons, and rather reflect genuine shortcomings.

\subsection{Target Conditions\label{sec:target_conditions}}

To establish an interpretive baseline for satisfactory performance, we define a set of target conditions. For association metrics ($r$, $\rho$, ICC) the target is statistical significance ($p < 0.05$); for error and bias metrics (MAE, Bias), the target is a failure to \emph{reject} the null hypothesis ($p\ge0.05$), indicating that the error is not statistically distinguishable from zero.
\section{Results}
\begin{table*}
    \centering
    \begin{threeparttable}
    \caption{Full-scale performance metrics for the ADAPTS framework, combined datasets ($N=204$).}
    \label{tab:global_performance}
        \begin{tabular}{ll|cccccc}
        \toprule
        \textbf{Instrument} & \textbf{Model} & \textbf{MAE} & \textbf{RMSE} & \textbf{Pearson's $r$} & \textbf{Spearman's $\rho$} & \textbf{ICC (3,1)} & \textbf{ICC (2,1)} \\ \midrule
         HAM-D 17* & DeepSeek R1 & 3.368 & 4.456 & 0.800 & 0.791 & 0.790 & 0.774 \\
                   & Gemini Pro & 3.319 & 4.353 & 0.841 & 0.824 & 0.822 & 0.781 \\
                   & GPT OSS & \textbf{3.216} & \textbf{4.177} & 0.811 & 0.801 & 0.799 & \textbf{0.795} \\
                   & Llama Scout 4 & 4.363 & 5.219 & 0.841 & \textbf{0.848} & \textbf{0.839} & 0.743 \\
                   & Claude Sonnet 4 & 3.588 & 4.662 & \textbf{0.852} & 0.841 & 0.829 & 0.755 \\
         \midrule
         HAM-D 17* (Extended) & DeepSeek R1 & 3.691 & 4.643 & \textbf{0.877} & \textbf{0.869} & \textbf{0.865} & 0.774 \\
                   & Gemini Pro & 3.833 & 4.870 & 0.859 & 0.848 & 0.836 & 0.741 \\
                   & GPT OSS & \textbf{2.740} & \textbf{3.670} & 0.854 & 0.843 & 0.844 & \textbf{0.844} \\
                   & Llama Scout 4 & 3.373 & 4.306 & 0.848 & 0.857 & 0.846 & 0.810 \\
                   & Claude Sonnet 4 & 3.892 & 5.024 & 0.856 & 0.843 & 0.830 & 0.724 \\
         \midrule
         HAM-A 14* & DeepSeek R1 & 2.414 & 3.204 & 0.922 & 0.919 & 0.922 & 0.919 \\
                   & Gemini Pro & \textbf{2.184} & \textbf{3.019} & \textbf{0.928} & \textbf{0.923} & \textbf{0.927} & \textbf{0.926} \\
                   & GPT OSS & 3.949 & 4.949 & 0.906 & 0.902 & 0.905 & 0.831 \\
                   & Llama Scout 4 & 16.640 & 18.057 & 0.619 & 0.589 & 0.617 & 0.209 \\
                   & Claude Sonnet 4 & 2.453 & 3.257 & 0.915 & 0.910 & 0.915 & 0.915 \\
         \bottomrule
    \end{tabular}
    \begin{tablenotes}
        \scriptsize
        \item Asterisks (*) denote modified text-evaluable subsets: HAM-D 17 excludes items 8 and 9 (Psychomotor Retardation/Agitation) and HAM-A 14 excludes item 14 (Behavior at Interview). The `Extended' HAM-D variant incorporates item-level qualitative clinical heuristics within model prompts for ambiguous symptomatic presentations.
    \end{tablenotes}
    \end{threeparttable}
\end{table*}
\begin{figure}
    \centering
    \resizebox{0.9\columnwidth}{!}{\input{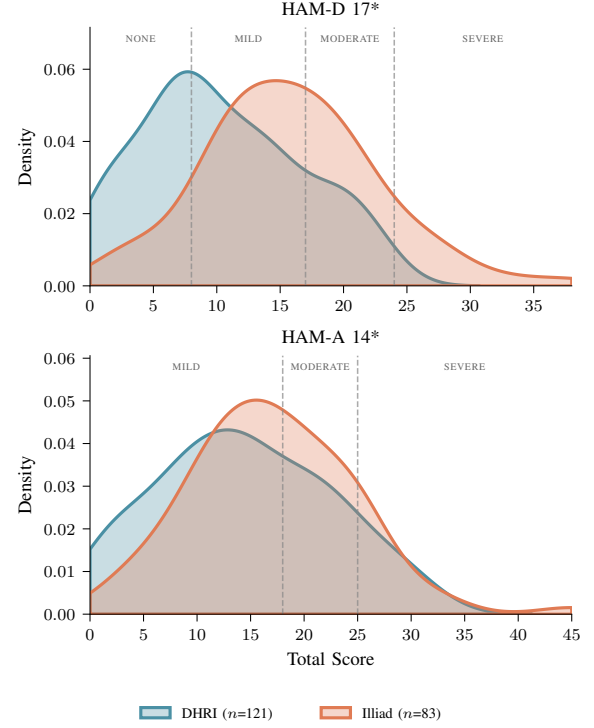}}
    \caption{Distribution of expert ground truth total scores across datasets.}
    \label{fig:score_distribution}
\end{figure}

The full-scale performance metrics, presented in Table~\ref{tab:global_performance} for the modified scales (HAM-D 17* and HAM-A 14*), evaluate the LLM agents' ability to synthesize an overall clinical impression from interview transcripts. As illustrated in Figure~\ref{fig:score_distribution}, the combined datasets represent a broad spectrum of clinical distress, spanning sub-clinical symptoms to severe pathology. Across both instruments and all five models tested, the modular framework consistently met the target conditions for association and consistency, with significant $p$-values for Pearson's $r$, Spearman's $\rho$, and the Intraclass Correlation Coefficient (ICC(3,1)). This indicates that the modular framework reliably captures the relative symptom severity across the patient population, effectively replicating the expert's rank-ordering of cases regardless of the underlying model architecture.

However, a notable finding also apparent is the divergence between Mean Absolute Error (MAE) and Root Mean Square Error (RMSE). While high-parameter models like Gemini Pro and Claude Sonnet 4 frequently satisfied the target condition for RMSE ($p \ge 0.05$), they consistently failed to meet the target for MAE ($p < 0.05$). Statistically, this pattern indicates that while the agents' ratings are characterized by a significant systematic bias --- a consistent point-offset relative to the expert --- they are not prone to the extreme outliers or spurious ``hallucinations'' that would typically inflate the RMSE. This suggests that the LLM agents using a zero-shot approach are stable but not calibrated with the rating anchors of the human raters in these datasets: they perceive the clinical signal of distress but apply a different internal severity threshold than the human raters.

At the global level, Gemini Pro and Claude Sonnet 4 emerged as the most robust raters. For the standard HAM-D 17*, Sonnet 4 demonstrated the highest level of concordance ($r = 0.852$, $\rho = 0.841$). On the HAM-A 14*, Gemini Pro achieved the optimal performance, maintaining the lowest average error magnitude ($\text{MAE} = 2.184$) and highest absolute agreement ($\text{ICC(2,1)} = 0.926$).

\subsection{Impact of Clinical Qualitative Conventions}

The transition from the standard HAM-D 17* to the ``Extended'' protocol, which augments the modular prompts with qualitative clinical conventions, such as the exclusion of pre-existing chronic symptoms or external environmental factors, yielded marked improvements in relative association and consistency. As shown in Table~\ref{tab:global_performance}, the inclusion of these expert-level guidelines stabilized the ratings for several models. DeepSeek R1, for instance, saw a significant increase in relative agreement, with Pearson's correlation rising from $r=0.800$ to $r=0.877$ and $\text{ICC(3,1)}=0.790$ rising to $\text{ICC(3,1)}=0.865$. While absolute error metrics slightly increased for proprietary models under this protocol, GPT OSS demonstrated marked gains in calibration, with a reduction from $\text{MAE}=3.216$ to $\text{MAE}=2.740$ and absolute agreement $\text{ICC(2,1)}=0.795$ increasing to $\text{ICC(2,1)}=0.844$. These results suggest that providing LLMs with explicit normative anchors can mitigate the subjective hypersensitivity previously observed in zero-shot clinical inference.

\subsection{Item-Level Rating Reliability and Bias Direction}
\begin{table*}[t]
\centering
\begin{threeparttable}
    \caption{Item-level performance metrics for the ADAPTS framework, combined datasets ($N=204$).}
    \label{tab:item_performance}
        \begin{tabular}{ll|cc|cc|cc|cc|cc}
        \toprule
        &  & \multicolumn{2}{c}{\textbf{Claude Sonnet 4}} & \multicolumn{2}{c}{\textbf{Llama Scout 4}} & \multicolumn{2}{c}{\textbf{DeepSeek R1}} & \multicolumn{2}{c}{\textbf{GPT OSS}} & \multicolumn{2}{c}{\textbf{Gemini Pro}} \\
        \textbf{{Item}} & \textbf{{Symptom}} & \textbf{{MAE}} & \textbf{{Pearson $r$}} & \textbf{{MAE}} & \textbf{{Pearson $r$}} & \textbf{{MAE}} & \textbf{{Pearson $r$}} & \textbf{{MAE}} & \textbf{{Pearson $r$}} & \textbf{{MAE}} & \textbf{{Pearson $r$}} \\ \midrule
        \multicolumn{2}{l}{HAM-D 17*} & \multicolumn{10}{c}{} \\
        \quad  1 & Depressed Mood & \textcolor{gray}{\itshape 0.471} & 0.752 & \textcolor{gray}{\itshape 0.833} & 0.700 & 0.534 & 0.711 & \textcolor{gray}{\itshape 0.706} & \textbf{0.746} & 0.466 & 0.753 \\
        \quad  2 & Guilt & 0.525 & 0.712 & \textcolor{gray}{\itshape 0.627} & 0.712 & \textcolor{gray}{\itshape 0.505} & 0.768 & \textcolor{gray}{\itshape 0.676} & 0.718 & \textcolor{gray}{\itshape 0.453} & 0.764 \\
        \quad  3 & Suicide & \textbf{0.154} & \textbf{0.833} & \textcolor{gray}{\itshape 0.208} & \textbf{0.776} & \textbf{0.135} & \textbf{0.853} & \textcolor{gray}{\itshape 0.257} & 0.709 & \textbf{0.115} & \textbf{0.866} \\
        \quad  4 & Insomnia Early & \textcolor{gray}{\itshape 0.284} & 0.747 & \textcolor{gray}{\itshape 0.775} & 0.493 & \textcolor{gray}{\itshape 0.284} & 0.766 & \textcolor{gray}{\itshape 0.353} & 0.714 & \textcolor{gray}{\itshape 0.304} & 0.763 \\
        \quad  5 & Insomnia Middle & 0.255 & 0.750 & \textcolor{gray}{\itshape 0.848} & 0.353 & \textcolor{gray}{\itshape 0.294} & 0.702 & \textbf{0.436} & 0.528 & 0.304 & 0.711 \\
        \quad  6 & Insomnia Late & 0.174 & 0.759 & \textcolor{gray}{\itshape 0.949} & 0.345 & 0.199 & 0.734 & \textcolor{gray}{\itshape 0.306} & 0.593 & 0.174 & 0.768 \\
        \quad  7 & Work/Activities & \textcolor{gray}{\itshape 0.507} & 0.762 & \textcolor{gray}{\itshape 1.184} & 0.646 & \textcolor{gray}{\itshape 0.615} & 0.725 & \textcolor{gray}{\itshape 0.811} & 0.691 & \textcolor{gray}{\itshape 0.586} & 0.716 \\
        \quad 10 & Anxiety Psychic & \textcolor{gray}{\itshape 1.054} & 0.317 & \textcolor{gray}{\itshape 0.902} & 0.366 & \textcolor{gray}{\itshape 1.005} & 0.260 & \textcolor{gray}{\itshape 1.020} & 0.221 & \textcolor{gray}{\itshape 1.113} & 0.202 \\
        \quad 11 & Anxiety Somatic & \textcolor{gray}{\itshape 0.917} & 0.313 & \textcolor{gray}{\itshape 1.113} & 0.344 & \textcolor{gray}{\itshape 0.868} & 0.308 & \textcolor{gray}{\itshape 1.005} & 0.330 & \textcolor{gray}{\itshape 0.784} & 0.373 \\
        \quad 12 & Somatic GI & \multicolumn{1}{c}{---} & \multicolumn{1}{c|}{---} & \textcolor{gray}{\itshape 0.551} & 0.250 & \textcolor{gray}{\itshape 0.571} & \textcolor{gray}{\itshape 0.025} & \multicolumn{1}{c}{---} & \multicolumn{1}{c|}{---} & \multicolumn{1}{c}{---} & \multicolumn{1}{c}{---} \\
        \quad 13 & Somatic General & \textcolor{gray}{\itshape 0.779} & 0.324 & \textcolor{gray}{\itshape 0.608} & 0.358 & \textcolor{gray}{\itshape 0.784} & 0.317 & \textcolor{gray}{\itshape 0.662} & 0.354 & \textcolor{gray}{\itshape 0.740} & 0.337 \\
        \quad 14 & Genital Symptoms & \textcolor{gray}{\itshape 0.525} & \textcolor{gray}{\itshape 0.112} & \textbf{0.593} & \textcolor{gray}{\itshape 0.109} & \textcolor{gray}{\itshape 0.480} & 0.172 & 0.618 & \textcolor{gray}{\itshape 0.089} & \textcolor{gray}{\itshape 0.522} & \textcolor{gray}{\itshape 0.094} \\
        \quad 17 & Insight & \multicolumn{1}{c}{---} & \multicolumn{1}{c|}{---} & \textcolor{gray}{\itshape 0.270} & 0.272 & 0.167 & 0.246 & \textcolor{gray}{\itshape 0.211} & 0.230 & \textcolor{gray}{\itshape 0.088} & \textcolor{gray}{\itshape 0.094} \\
        \midrule
        \multicolumn{2}{l}{HAM-A 14*} & \multicolumn{10}{c}{} \\
        \quad  1 & Anxious Mood & 0.387 & 0.760 & \textcolor{gray}{\itshape 2.015} & \textcolor{gray}{\itshape 0.023} & 0.436 & 0.703 & \textcolor{gray}{\itshape 0.559} & 0.663 & 0.407 & 0.762 \\
        \quad  2 & Tension & \textcolor{gray}{\itshape 0.395} & 0.739 & \textcolor{gray}{\itshape 2.233} & 0.152 & 0.350 & 0.786 & \textcolor{gray}{\itshape 0.414} & 0.750 & 0.331 & 0.781 \\
        \quad  3 & Fears & \textcolor{gray}{\itshape 0.314} & 0.833 & \textcolor{gray}{\itshape 1.230} & 0.495 & \textcolor{gray}{\itshape 0.338} & 0.818 & \textcolor{gray}{\itshape 0.534} & 0.758 & \textcolor{gray}{\itshape 0.328} & 0.806 \\
        \quad  4 & Insomnia & \textcolor{gray}{\itshape 0.525} & 0.742 & \textcolor{gray}{\itshape 1.730} & 0.273 & \textcolor{gray}{\itshape 0.578} & 0.687 & \textcolor{gray}{\itshape 0.745} & 0.590 & \textcolor{gray}{\itshape 0.578} & 0.680 \\
        \quad  5 & Intellectual & \textcolor{gray}{\itshape 0.417} & 0.817 & \textcolor{gray}{\itshape 0.750} & 0.577 & 0.382 & 0.808 & \textcolor{gray}{\itshape 0.554} & 0.716 & 0.368 & 0.808 \\
        \quad  6 & Depressed Mood & \textcolor{gray}{\itshape 0.635} & 0.594 & \textcolor{gray}{\itshape 2.091} & \textcolor{gray}{\itshape 0.121} & 0.512 & 0.713 & \textcolor{gray}{\itshape 0.870} & 0.528 & 0.404 & 0.790 \\
        \quad  7 & Somatic (Muscular) & 0.341 & 0.791 & \textcolor{gray}{\itshape 1.336} & 0.430 & \textcolor{gray}{\itshape 0.306} & 0.822 & \textcolor{gray}{\itshape 0.365} & 0.786 & 0.282 & 0.819 \\
        \quad  8 & Somatic (Sensory) & 0.275 & 0.814 & \textcolor{gray}{\itshape 1.873} & 0.264 & \textbf{0.230} & 0.842 & \textcolor{gray}{\itshape 0.353} & 0.797 & 0.211 & 0.837 \\
        \quad  9 & Cardiovascular & \textcolor{gray}{\itshape 0.262} & 0.796 & \textcolor{gray}{\itshape 1.179} & 0.399 & 0.262 & 0.786 & \textcolor{gray}{\itshape 0.341} & 0.765 & 0.218 & 0.825 \\
        \quad 10 & Respiratory & \textcolor{gray}{\itshape 0.218} & \textbf{0.851} & \textcolor{gray}{\itshape 0.816} & 0.391 & 0.238 & 0.805 & \textcolor{gray}{\itshape 0.248} & \textbf{0.863} & \textbf{0.184} & 0.826 \\
        \quad 11 & Gastrointestinal & \textbf{0.255} & 0.850 & \textcolor{gray}{\itshape 1.505} & 0.242 & \textbf{0.230} & \textbf{0.871} & \textcolor{gray}{\itshape 0.431} & 0.761 & 0.216 & \textbf{0.876} \\
        \quad 12 & Genitourinary & 0.257 & 0.812 & \textcolor{gray}{\itshape 0.517} & \textbf{0.697} & 0.243 & 0.788 & \textcolor{gray}{\itshape 0.355} & 0.741 & 0.267 & 0.766 \\
        \quad 13 & Autonomic & \textcolor{gray}{\itshape 0.277} & 0.817 & \textcolor{gray}{\itshape 2.458} & 0.251 & 0.306 & 0.795 & \textcolor{gray}{\itshape 0.404} & 0.792 & 0.233 & 0.838 \\
        \bottomrule
    \end{tabular}
    \begin{tablenotes}
        \scriptsize
        \item Asterisks (*) denote modified text-evaluable subsets: HAM-D 17 excludes items 8 and 9 (Psychomotor Retardation/Agitation) and HAM-A 14 excludes item 14 (Behavior at Interview). \textbf{Boldface} indicates optimal results among those meeting target conditions ($p < 0.05$ for association; $p \ge 0.05$ for bias); failures are denoted in \textit{\color{gray}grey italics}. All $p$-values are Benjamini-Hochberg adjusted. See Section~\ref{sec:target_conditions} (page~\pageref{sec:target_conditions}) for further detail on target conditions.
    \end{tablenotes}
\end{threeparttable}
\end{table*}

Item-level analysis reveals that the direction and magnitude of bias are highly symptom-dependent, even across different model families.

\subsubsection{HAM-D 17* Items}

As detailed in Table~\ref{tab:item_performance}, high-parameter models functioned as reliable raters for core depressive symptoms. Gemini Pro and DeepSeek R1 satisfied association target conditions for Depressed Mood, Guilt, and Suicide. Notably, Gemini Pro reached near-human accuracy on the Suicide item, achieving the highest concordance ($r = 0.866$) and the lowest error magnitude ($\text{MAE} = 0.115$). In these core domains, the reasoning-focused models achieved near-zero bias (e.g., Gemini Pro Guilt $\text{Bias} = -0.054$), meaning their scores were statistically indistinguishable from human rating labels.

Conversely, all models exhibited a systematic tendency toward over-estimation when rating the Insomnia cluster (Items 4--6) and Anxiety symptoms (Items 10--11). In these domains, the residuals were statistically distinguishable from zero ($p < 0.05$) and mostly positive. This confirms that while models can identify the presence of these symptoms, they struggle with the calibration layer required to distinguish transient linguistic distress from persistent clinical pathology without the benefit of the Extended protocol's conventions.

\subsubsection{HAM-A 14* Items}

On the HAM-A scale, Gemini Pro and Claude Sonnet 4 met the target conditions for a broader range of symptoms, including Anxious Mood, Tension, and various somatic markers. Gemini Pro provided particularly accurate ratings for Gastrointestinal symptoms, yielding a Pearson's $r$ of $0.876$ and a low average error ($\text{MAE} = 0.216$).

While Llama Scout 4 generally struggled with systematic bias, frequently over-scoring items like Insomnia ($\text{MAE} = 1.730$), it maintained rank-order association for specific markers like Genitourinary symptoms ($r=0.697$). Consistent with the HAM-D results, the agents struggled to generate unbiased scores for complex Autonomic symptoms, where the magnitude of the error remained statistically distinguishable from expert labels across all benchmarked architectures.

\begin{figure*}
    \centering
    \resizebox{0.9\textwidth}{!}{\input{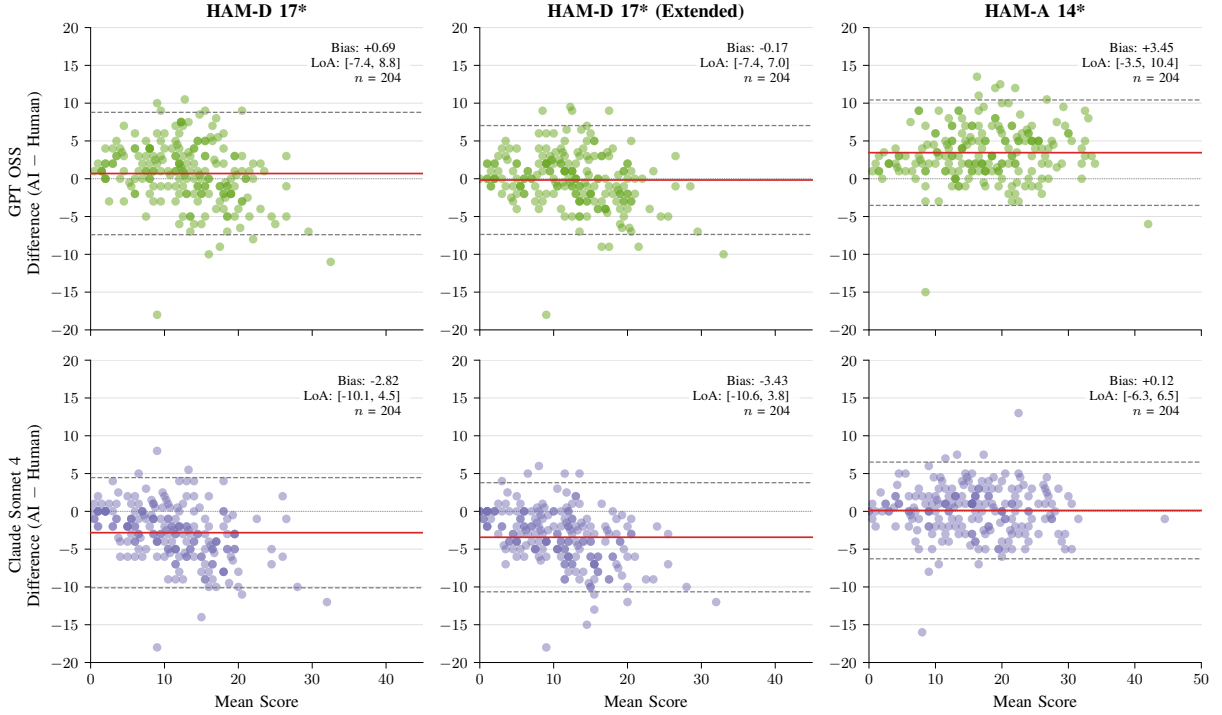}}
    \caption{Bland-Altman plots evaluating agreement between LLM ratings and expert benchmarks for Claude Sonnet 4.5 and GPT OSS. Solid lines indicate mean bias (point-offset); dashed lines represent 95\% limits of agreement ($\pm 1.96$ SD).}
\end{figure*}

\subsection{Dataset Variance and Generalizability}
\begin{table*}
    \centering
    \begin{threeparttable}
    \caption{Full-scale performance metrics for the ADAPTS framework, disaggregated by dataset (DHRI, $n=121$; Illiad, $n=83$)}
    \label{tab:dataset_performance}
        \begin{tabular}{lll|cccccc}
        \toprule
        \textbf{Instrument} & \textbf{Dataset} & \textbf{Model} & \textbf{MAE} & \textbf{RMSE} & \textbf{Pearson's $r$} & \textbf{Spearman's $\rho$} & \textbf{ICC (3,1)} & \textbf{ICC (2,1)} \\ \midrule
         HAM-D 17* & DHRI & DeepSeek R1 & 2.678 & 3.585 & 0.829 & 0.849 & 0.828 & 0.829 \\
                   &                 & Gemini 3 Pro & \textbf{2.248} & \textbf{2.839} & \textbf{0.895} & \textbf{0.902} & \textbf{0.890} & \textbf{0.888} \\
                   &                 & GPT OSS & 3.140 & 3.926 & 0.858 & 0.871 & 0.857 & 0.807 \\
                   &                 & Llama Scout 4 & 4.950 & 5.693 & 0.886 & 0.899 & 0.883 & 0.699 \\
                   &                 & Claude Sonnet 4 & 2.694 & 3.626 & 0.860 & 0.876 & 0.847 & 0.812 \\
                   & Illiad & DeepSeek R1 & 4.373 & 5.483 & 0.791 & 0.764 & 0.783 & 0.681 \\
                   &                 & Gemini 3 Pro & 4.880 & 5.902 & 0.819 & 0.770 & 0.804 & 0.645 \\
                   &                 & GPT OSS & \textbf{3.325} & 4.518 & 0.782 & 0.730 & 0.768 & 0.749 \\
                   &                 & Llama Scout 4 & 3.506 & \textbf{4.438} & 0.804 & \textbf{0.802} & 0.800 & \textbf{0.776} \\
                   &                 & Claude Sonnet 4 & 4.892 & 5.852 & \textbf{0.829} & 0.791 & \textbf{0.812} & 0.651 \\
         \midrule
         HAM-D 17* (Extended) & DHRI & DeepSeek R1 & 2.702 & 3.388 & \textbf{0.904} & \textbf{0.911} & \textbf{0.899} & 0.849 \\
                   &                 & Gemini 3 Pro & 2.512 & 3.230 & 0.898 & \textbf{0.911} & 0.891 & 0.857 \\
                   &                 & GPT OSS & \textbf{2.339} & \textbf{3.137} & 0.881 & 0.891 & 0.879 & \textbf{0.870} \\
                   &                 & Llama Scout 4 & 3.620 & 4.503 & 0.880 & 0.904 & 0.879 & 0.784 \\
                   &                 & Claude Sonnet 4 & 2.884 & 3.933 & 0.864 & 0.881 & 0.849 & 0.785 \\
                   & Illiad & DeepSeek R1 & 5.133 & 6.020 & 0.844 & 0.801 & \textbf{0.837} & 0.659 \\
                   &                 & Gemini 3 Pro & 5.759 & 6.564 & \textbf{0.855} & 0.812 & 0.835 & 0.600 \\
                   &                 & GPT OSS & 3.325 & 4.330 & 0.817 & 0.784 & 0.807 & 0.779 \\
                   &                 & Llama Scout 4 & \textbf{3.012} & \textbf{4.000} & 0.818 & \textbf{0.828} & 0.815 & \textbf{0.816} \\
                   &                 & Claude Sonnet 4 & 5.361 & 6.283 & 0.830 & 0.771 & 0.810 & 0.613 \\
         \midrule
         HAM-A 14* & DHRI & DeepSeek R1 & 2.211 & 2.861 & 0.938 & 0.937 & 0.938 & 0.938 \\
                   &                 & Gemini 3 Pro & \textbf{1.921} & \textbf{2.647} & \textbf{0.946} & \textbf{0.946} & \textbf{0.946} & \textbf{0.946} \\
                   &                 & GPT OSS & 4.293 & 5.262 & 0.923 & 0.920 & 0.919 & 0.826 \\
                   &                 & Llama Scout 4 & 16.855 & 18.069 & 0.723 & 0.694 & 0.717 & 0.249 \\
                   &                 & Claude Sonnet 4 & 2.467 & 3.206 & 0.922 & 0.920 & 0.922 & 0.921 \\
                   & Illiad & DeepSeek R1 & 2.711 & 3.645 & 0.902 & 0.887 & 0.902 & 0.885 \\
                   &                 & Gemini 3 Pro & 2.566 & 3.490 & 0.900 & 0.889 & 0.899 & 0.889 \\
                   &                 & GPT OSS & 3.446 & 4.453 & 0.885 & 0.883 & 0.885 & 0.837 \\
                   &                 & Llama Scout 4 & 16.325 & 18.040 & 0.416 & 0.414 & 0.416 & 0.132 \\
                   &                 & Claude Sonnet 4 & \textbf{2.434} & \textbf{3.329} & \textbf{0.903} & \textbf{0.895} & \textbf{0.903} & \textbf{0.903} \\
         \bottomrule
    \end{tabular}
    \begin{tablenotes}
        \scriptsize
        \item Asterisks (*) denote modified text-evaluable subsets: HAM-D 17 excludes items 8 and 9 (Psychomotor Retardation/Agitation) and HAM-A 14 excludes item 14 (Behavior at Interview). The `Extended' HAM-D variant incorporates item-level qualitative clinical heuristics within model prompts for ambiguous symptomatic presentations.
    \end{tablenotes}
    \end{threeparttable}
\end{table*}

Disaggregating performance by dataset (Table~\ref{tab:dataset_performance}) confirms that the ADAPTS framework is robust to significant variations in interview protocol. Despite the structural differences between the collapsed DHRI protocol ($n=121$) and the sequential Illiad protocol ($n=83$), the top-tier models maintained high rank-order fidelity across both cohorts.

Notably, performance on the DHRI dataset consistently yielded higher association metrics; for the standard HAM-D 17*, Gemini Pro achieved an $\text{ICC(3,1)}$ of $0.890$ on DHRI compared to $0.804$ on Illiad. This divergence likely reflects the higher density of symptom-relevant content in the DHRI protocol's non-linear flow. Furthermore, the Extended HAM-D protocol proved effective in stabilizing variance across cohorts: DeepSeek R1 achieved an absolute agreement of $\text{ICC(2,1)} = 0.849$ on the DHRI sample while simultaneously maintaining a high consistency $\text{ICC(3,1)}$ of $0.837$ on the Illiad sample.

The consistency of these results across distinct datasets suggests that the ADAPTS framework's retrieval mechanism effectively mitigates ``timestamp drift'' and chronological dependencies. By treating the interview as an agnostic evidence pool rather than a linear questionnaire, the sub-agents successfully localized symptomatic data regardless of whether the prompts were sequential or interwoven. This generalizability is a necessary prerequisite for deployment in real-world clinical environments where semi-structured dialogue often deviates from rigid instrument sequences.
\section{Discussion}

The present study examined whether large language models can reliably and meaningfully re-rate clinician-administered assessments of depression and anxiety from naturalistic clinical interviews. Specifically, we tested whether a modular, item-level LLM architecture could recover expert rank-orderings of symptom severity across heterogeneous interview protocols, mitigate known long-context reasoning failures in LLMs~\cite{liu2024lost}, and disentangle stable model behavior from systematic calibration bias, an issue often obscured when evaluation relies primarily on correlation metrics [12], [13]. Across two independent datasets with distinct interview structures, LLM-based ratings demonstrated strong consistency with expert severity rankings, as evidenced by significant Pearson, Spearman, and ICC(3,1) coefficients. At the same time, models exhibited persistent absolute error, reflected in significant mean absolute error but frequently non-significant RMSE, indicating stable yet miscalibrated scoring thresholds rather than hallucination-driven instability. Importantly, introducing explicit clinical conventions into the prompting framework improved absolute agreement, suggesting that calibration errors reflect missing normative constraints rather than fundamental representational limitations. Together, these findings support the feasibility of LLM-based psychiatric re-rating while clarifying the conditions required for clinical reliability and interpretability.

\subsection{Stability vs. Calibration}

The observed dissociation between rank-order reliability and absolute calibration reinforces longstanding concerns in both affective computing and clinical measurement that correlation alone is an insufficient criterion for evaluating AI-based mental health assessment systems~\cite{koo2016guideline}. Recent LLM-based studies of interview-derived symptom scoring have reported high correlations with clinician ratings~\cite{raganatoleveraging,kebe2025llamadrs}, yet correlation metrics remain insensitive to systematic bias. In the present study, models consistently replicated expert severity ordering while exhibiting stable point offsets, a pattern consistent with psychometric work showing that raters may be reliable but poorly calibrated~\cite{kobak2004rater,muller2002effects}. This finding underscores the importance of jointly reporting association, ICC-based reliability, and absolute error metrics when evaluating affective AI systems intended for clinical use.

\subsection{Modular Item-Level Reasoning}

The modular, item-level architecture adopted here addresses well-documented limitations of LLMs in long-context settings, including position effects and degraded evidence utilization when relevant information appears mid-document~\cite{liu2024lost}. Rather than prompting a single model to infer global severity from extended clinical dialogue, the framework decomposes assessment into symptom-specific agents that retrieve evidence across the entire transcript. This design aligns with clinician-administered instruments such as the HAM-D and HAM-A, which structure judgment at the item level rather than as a holistic impression~\cite{hamilton1959assessment,hamilton1960rating}. The strong generalization observed across distinct datasets suggests that architectural decomposition meaningfully mitigates long-context degradation and supports reliable affective inference in real-world clinical dialogue.

\subsection{Addressing Protocol Heterogeneity}

Many automated mental health assessment systems implicitly rely on stable interview structures or fixed question–answer sequences, often exploiting ordering regularities rather than symptom-relevant content~\cite{burdisso2024daicwoz}. However, clinical assessment protocols vary widely in practice, with scales administered sequentially, interwoven, or collapsed to reduce burden~\cite{demyttenaere2020trends}. The present framework was designed to retrieve symptom evidence irrespective of its position in the interview, enabling robust performance across both linear and non-linear protocols. These results align with recent findings showing that interviewer prompts and ordering artifacts can bias automated depression detection when treated as predictive features rather than contextual scaffolding~\cite{burdisso2024daicwoz}. Treating assessment as a retrieval-and-justification task rather than a positional mapping problem therefore represents a critical step toward protocol-robust affective AI.

\subsection{Calibration as a Knowledge-Injection Problem Rather Than Model Failure}

A central contribution of this work is the demonstration that LLM calibration errors can be attenuated through explicit injection of qualitative clinical conventions. Augmenting prompts with structured interpretive guidance, such as discounting externally driven sleep disturbance, substantially improved absolute agreement, particularly for reasoning-oriented models. This finding supports the view that LLMs encode rich representations of affective meaning but lack access to domain-specific normative thresholds required for standardized scoring~\cite{raganatoleveraging,weber2025using}. Rather than reflecting unreliability, the observed bias appears to arise from underspecified scoring conventions. These results suggest that principled knowledge injection may serve as a scalable calibration strategy, avoiding the need for task-specific fine-tuning on sensitive clinical data.

\subsection{Symptom-Specific Performance and the Limits of Text-Only Inference}

Consistent with prior affective computing research, model performance varied systematically by symptom domain. Semantically explicit constructs such as suicidality and guilt were inferred with near-human accuracy, whereas symptoms relying on physiological or behavioral observation, such as autonomic anxiety or psychomotor disturbance, showed persistent bias~\cite{hamilton1959assessment,hamilton1960rating}. This pattern reflects long-recognized distinctions in clinical assessment between verbally reportable symptoms and those requiring visual or paralinguistic cues. While the present results demonstrate the feasibility of text-based psychiatric re-rating, they also highlight the necessity of multimodal approaches that integrate speech prosody, facial behavior, and movement dynamics to capture the full range of clinically relevant affective signals~\cite{valstar2013avec,devault2014simsensei}.
\section{Limitations and Future Work}

Several limitations should be considered when interpreting these findings. First, the present implementation relies exclusively on transcribed text and therefore cannot directly assess symptoms that depend on non-verbal, paralinguistic, or behavioral cues. This limitation is most evident for items such as psychomotor agitation or retardation and observable anxious behavior, which were intentionally excluded from automated scoring. While this constraint represents a methodological safeguard against unsupported inference, it also underscores an important boundary condition: text-only LLMs are inherently limited in their ability to capture aspects of affective expression that clinicians routinely integrate through visual and acoustic observation. Future work should therefore extend this modular framework to multimodal inputs, incorporating speech prosody, facial behavior, and movement dynamics to support more comprehensive affective assessment.

Second, although the study evaluated generalization across two independent datasets with distinct interview protocols, both datasets were drawn from structured clinical research contexts and involved trained raters. Performance in less controlled clinical environments, such as routine outpatient care, crisis services, or telehealth settings, remains an open question. Additional validation across more diverse populations, languages, and clinical contexts will be essential to assess robustness, fairness, and real-world applicability.

Third, while the introduction of explicit clinical conventions substantially improved calibration, the present work does not claim to have identified an optimal or exhaustive set of such constraints. Calibration remains an iterative process, and future research should systematically examine which types of clinical knowledge (e.g., heuristics, anchoring rules) most effectively align LLM outputs with expert judgment. Importantly, this raises broader questions about how clinical norms should be encoded, governed, and updated in AI-assisted assessment systems.

Finally, the current framework was evaluated as an offline re-rating tool rather than as a real-time clinical support system. Future work should explore how modular LLM agents might be integrated into live or near real-time workflows, such as quality assurance, rater calibration, or clinician-in-the-loop screening. In such applications, careful attention will be required to ensure that AI-generated ratings augment rather than displace clinical judgment, and that transparency and accountability remain central design principles.

\section{Conclusion}

Taken together, these findings advance affective computing research by demonstrating that LLMs can function as stable, interpretable components of mental health measurement infrastructure when appropriately constrained and evaluated. Rather than serving as autonomous diagnosticians, modular LLM systems may be best positioned as assistive raters, supporting calibration, identifying inconsistencies, and stabilizing noisy endpoints in both research and practice. By explicitly distinguishing reliability from calibration and grounding inference in item-level reasoning aligned with clinical instruments, this work contributes the ADAPTS framework for deploying affective AI systems in safety-critical mental health contexts. The study limitations point toward a broader research agenda in which LLM-based affective systems are developed not as standalone diagnostic authorities, but as modular, multimodal, and clinically grounded tools that support reliable measurement while respecting the complexity of human judgment in mental health care.

\bibliographystyle{IEEEtran}
\bibliography{taffc}

\vfill

\end{document}